\documentclass[runningheads]{llncs}
\usepackage[T1]{fontenc}
\usepackage{amssymb}
\usepackage{graphicx}
\usepackage{color}
\begin{document}
\title{Hallucinations and constraints : Regulating surgical workflow recognition beyond accuracy}
\titlerunning{Hallucinations and constraints}
%
\author{John S.H. Baxter\orcidID{0000-0003-3548-4343} \and Pierre Jannin\orcidID{0000-0002-7415-071X}}

\authorrunning{J.S.H. Baxter \& P. Jannin}
%
\institute{Anonymous Affiliation}
\institute{Université de Rennes, Inserm, Laboratoire Traitement du Signal et de l'Image (LTSI - UMR 1099), F-35000 Rennes, France \email{john.baxter@univ-rennes.fr}}
\maketitle
\begin{abstract}
Hallucinations are a major concern for the integration of artificial intelligence into medicine, although less explored in the realm of medical image processing. Unlike problems in natural text understanding and reasoning therewith, determining whether or not predictions derived from biomedical images and signals is less intuitively clear. This article suggests that topological errors could constitute hallucinations in a way that can be more readily measured and thus regulated. Certain of these properties for certain types of problems, such as biomedical signal segmentation, can be rephrased as linear temporal logic predicates, a number of which can be explicitly enforced using probabilistic graphical models. Our simulations show the potential of these explicitly constrained predicates for the case of automatic surgical phase recognition in robot-assisted hysterectomy, improving accuracy by approximately 10\% while removing the vast majority of topological errors, suggesting that mathematical guarantees of correctness can supplement other empirical forms of regulating machine learning in medical image computing and computer-assisted interventions.

\end{abstract}

\section{Introduction}
Although currently associated with large language models (LLMs), the concept of \textit{hallucination} as a pejorative actually predates them with authors such as Rohrbach \textit{et al.} \cite{rohrbach2018object} coining the term \textit{object hallucination} for errors in image captioning in which the machine learning model would incorrectly state that a particular object appeared in an image. The more commonly discussed notion of hallucination, i.e. reporting factually incorrect statements from a corpus of text, was already of concern to the deep learning community as early as 2020 \cite{maynez2020faithfulness}, again before the proliferation of large language models in the form of ChatGPT's public release in 2022.

For these problems, the notion of a hallucination is relatively clear (it takes the form of an obvious factual error) and thus there has been no widespread discussion of what does or does not distinguish between a hallucination and an error. For medical imaging and computer-assisted interventions, a taxonomy of errors seems particularly important as we intuitively know that there are some errors that can be considered majors even if, quantitatively, they have the same value as minor ones. For example, if we imagine a binary brain tumour segmentation algorithm with 90\% Dice, that could represent either state-of-the-art performance (if those errors are on the sides of tumours) or completely unusable (if those errors take the form of additional tumours or missing metastases). 

In terms of regulating AI algorithms, having a clear distinction between non-hallucinations and hallucinations would be of the upmost importance for ensuring their safe use. As different medical imaging problems (especially those in different medical domains) have particular clinical requirements, what constitutes a hallucination should be specific to that particular problem. However, there may be families of errors that can commonly be interpreted as hallucinations across various domains, especially if they are technically similar.

The goal of this article is (i) to motivate one of these families, \textit{linear topoligical errors}, in the context of surgical workflow analysis, (ii) to show how such errors can be defined for a specific application, \textit{robot-assisted hysterectomy}, and (iii) to propose a new approach to addressing said errors that illustrates a potential new path for strong medical AI regulation: \textit{constrained AI}.

\section{Theory}
\subsection{Signal segmentation and linear topological predicates}
Signal segmentation is the application of semantic segmentation to data that is fundamentally indexed by a single dimension, often time. Mathematically, this can be construed as learning a function $F : (\mathbb{R}\to \mathbb{I})\to(\mathbb{R}\to 2^\mathbb{L})$ where $\mathbf{R^1}$ is the ``temporal'' dimension, $\mathbb{I}$ is whatever range space the signal has (i.e. $\mathbb{R}^1$ for scalar signals such as audio, or $(\mathbb{R}^2\to\mathbb{R}^3)$ for video, etc.), and $\mathbb{L}$ is a finite set of labels. The core idea is that each element of time from the original domain should be given zero or more labels by the model which represent what was happening at that particular instant. In computer-assisted interventions, these labels could be various things such as (i) what tools (if any) are visible at that instant in a surgical video, (ii) what anatomies are visible, or (iii) what the particular surgical task being performed is. The remainder of this article will focus on the latter, commonly known as \textit{surgical workflow recognition} \cite{lalys2014surgical,liu2025deep}.

For many surgeries, there are rules about the ways in which labels (i.e. surgical phases, steps, or actions) must appear. These rules can either be physical (e.g. each hand can hold at most one tool, etc.) logical (e.g. before a tumour can be removed, it must first be separated from the healthy anatomy) or procedural (e.g. it is best to check for functional deficits before removing pathology). At an abstract level, these rules (in surgery as well as other domains) often fall into a number of general classes that can be described via logical predicates \cite{xu2022don}.
Some of these \textit{linear topological predicates} include:
\begin{itemize}
    \item \underline{$A$ ``IMPLIES'' $B$}: Any instance labelled as $A$ is also labelled as $B$. (That is, $B$ is a group of labels that includes $A$ either as a sub-group or as a basic label.)
    \item \underline{$A$ ``PRECEDES'' $B$} : if $B$ appears in a signal then immediately before any block of $B$ time points must be an $A$ time point ($A$ ``NOT PRECEDES'' $B$ means the opposite, that the preceding time point is not labelled as $A$); and
    \item \underline{$A$ ``SUCCEEDS'' $B$} : if $B$ appears in a signal then immediately after any block of $B$ time points must be an $A$ time point ($A$ ``NOT SUCCEEDS'' $B$ again means the opposite in a similar way).
    \item \underline{$A$ ``IS INITIAL''}: No labels other than those of $A$ can appear before any from $A$.
    \item \underline{$A$ ``IS FINAL''}: No labels other than those of $A$ can appear after any from $A$.
\end{itemize}
amongst others. From these more simple predicates, other, more complex rules can be formed that capture notions of structured uncertainty, largely by grouping together labels using ``IMPLIES'' and then using the super-label with other predicates. The key similarity between these linear topological predicates is that they capture notions of potential orderings of phases at a more abstract level irrespective of their exact lengths and positions, but can be used when trying to classify whether or not a phase, step, or activity is occurring at a particular time $t$. These linear topological predicates also give us a definite class of errors that we can consider hallucinations: ones that break the fundamental ordering rules of the domain.

\subsection{Generic surgical process model for hysterectomy}
\begin{figure}[t!]
    \centering
    \includegraphics[height=0.35\textheight]{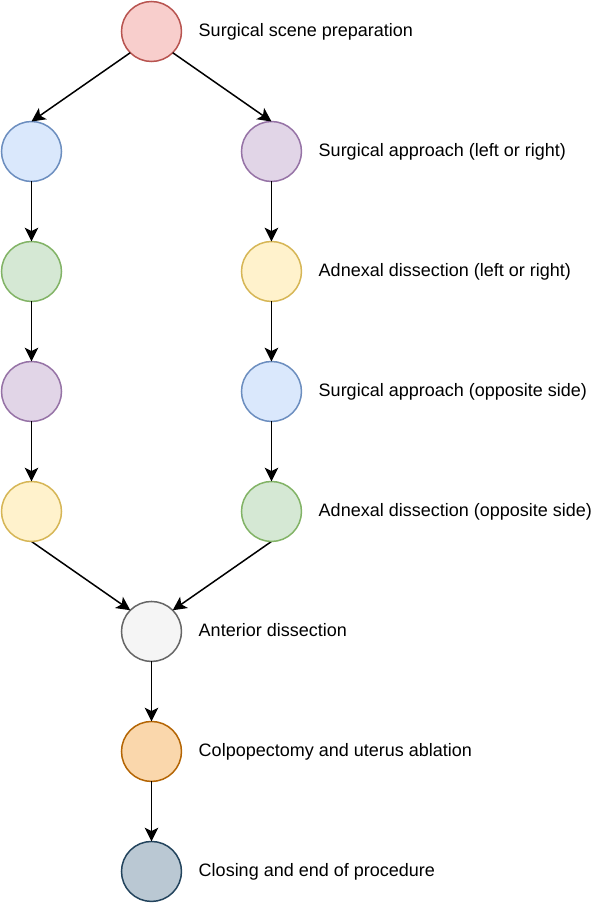}
    \vspace{-4mm}\caption{Simplified normative generic surgical process model for robot-assisted hysterectomy of a basic patient. Colours indicate when a phase is repeated in the graph, that is, the ability for the same surgical tasks to be performed validly in different orders.}
    \label{fig:gSPM}
\end{figure}

In the context of surgery, the linear temporal predicates can be captured through what are often called \textit{generic surgical process models} (gSPMs) which are finite state machines that say which blocks of actions can be performed after which other blocks, often at different levels of granularity \cite{lalys2014surgical,liu2025deep,nyangoh2026international}. (Individual surgical process models or \textit{iSPMs} are the list of said blocks as they were actually performed for an individual surgery.) These models can either be \textit{normative}, illustrating the space of `good' or `standard' procedures, or \textit{descriptive}, illustrating the space of all reasonable procedures. In either case, the gold standard for constructing these models is via a multi-expert Delphi consensus where a large number of experienced surgeons work together to create a gSPM that captures all the variability that they have encountered in their practice or expect to encounter, while being as precise and discriminating as possible \cite{nyangoh2026international}. This exercise is fundamentally important for regulating surgical workflow recognition models as it creates a standardised set of labels as well as a large collection of applicable linear topological predicates that should be followed. Figure \ref{fig:gSPM} displays a simplified surgical-phase-level gSPM for robot-assisted hysterectomy on a basic patient. This diagram captures the basic ordering of the most granular steps in a regular hysterectomy surgery, although oversimplified. (In reality, aspects of the anterior dissections can be performed before and between the adnexal dissections \cite{nyangoh2026international}.)

\newpage 
From this visual description as well as introducing some labels for grouping phases (e.g. \textit{Left-side phases} containing both the surgical approach and the adnexal dissection on the left side), a large number linear topological predicates can be extracted that all paths through Figure \ref{fig:gSPM} must adhere to, including:
\begin{enumerate}
    \item \textit{Surgical scene preparation} IS INITIAL
    \item \textit{Surgical approach (left)} PRECEDES \textit{Adnexal dissection (left)}
    \item  \textit{Adnexal dissection (left) }SUCEEDS \textit{Surgical approach (left)}
    \item \textit{Surgical approach (left)} IMPLIES \textit{Left-side phases}
    \item \textit{Adnexal dissection (left)} IMPLIES \textit{Left-side phases}
    \item \textit{Surgical approach (right)} PRECEDES \textit{Adnexal dissection (right)}
    \item  \textit{Adnexal dissection (right) }SUCEEDS \textit{Surgical approach (right)}
    \item \textit{Surgical approach (right)} IMPLIES \textit{Right-side phases}
    \item \textit{Adnexal dissection (right)} IMPLIES \textit{Right-side phases}
    \item \textit{Left-side phases} IMPLIES \textit{Lateral phases}
    \item \textit{Right-side phases} IMPLIES \textit{Lateral phases}
    \item \textit{Surgical scene preparation} PRECEDES \textit{Lateral phases}
    \item \textit{Lateral phases} SUCEEDS \textit{Surgical scene preparation}
    \item \textit{Lateral phases} PRECEDES \textit{Anterior dissection}
    \item \textit{Anterior dissection} SUCEEDS \textit{Lateral phases}
    \item \textit{Anterior dissection} PRECEDES \textit{Colpotomy and uterus ablation}
    \item \textit{Colpotomy and uterus ablation} SUCEEDS \textit{Anterior dissection}
    \item \textit{Colpotomy and uterus ablation} PRECEDES \textit{Closing and end of procedure}
    \item \textit{Closing and end of procedure} SUCEEDS \textit{Colpotomy and uterus ablation}
    \item \textit{Closing and end of procedure} IS FINAL
\end{enumerate}
Breaking any of these rules (such as predicting that the \textit{Closing and end of procedure} occurred at a particular time, but some \textit{Anterior dissection} occurred at a later time or that any lateral dissection was performed prior to the corresponding surgical approach) could be considered not just an error, but a hallucination. These are not the only topological rules, as even more rules could be determined that give a more descriptive list of potential hallucinations as well as more topological structure for models to use as prior knowledge. An additional measurable constraint not included in the model is uniqueness. That is, for each category in the gSPM, all instances occur in one contiguous block with no interruptions.

\subsection{Probabilistic graphical models and linear topological constraints}
Now that linear topological errors have been introduced and the construction of a particular set of said errors specific to robot-assisted hysterectomy have been formulated, the question now becomes how to integrate these together. One way to address this problem is to integrate the quantifications of these errors into different terms in the loss function used to train the machine learning model \cite{xu2022don,tayupo2026please}. The underlying motivation of this approach is that the machine learning model should, by virtue of being otherwise penalised, internalise the explicit topological rules encoded in the loss function. However, this internalisation is not always perfect \cite{tayupo2026please} and its degree may also depend on the amount of data, as more training data would allow for the algorithm to make more and different mistakes in training time which help it to learn the general meaning of the topological rule rather than only a list of particular instances. The general downside of this approach is that it is still susceptible to a form of the \textit{alignment problem}, that is, given a model large and flexible enough to solve a particular problem, it is large and flexible enough to have figured out a way to get around particular rules rather than be constrained by them.

Probabilistic graphical models (PGMs) can give us a tool to implement these constraints in the final activation layer of a deep learning model, replacing the traditional softmax/argmax layers with custom operations, specifically sum-product belief propagation \cite{pearl1987distributed} to replace softmax and either max-product belief propagation \cite{pearl1987distributed} or max-flow \cite{baxter2017directed,baxter2015shape} to replace argmax, all of which are well-understood and mathematically validated approaches with correctness guarantees for acyclic graphs.
Certain linear topological constraints can be easily integrated into an acyclic PGM where the graph in question is linear, simply representing temporal adjacency. This would allow certain predicates to be encoded if they only require knowledge of two adjacent time points to provide an example of their breakage. For example, if one has the predicate $A$ PRECEDES $B$ and one sees a string $AAABBABB$, then one can tell this predicate was broken simply due to the existence of substring $BA$. For our list of predicates, a number of them (IMPLIES, PRECEDES and SUCEEDS with their inverses, as well as IS INITIAL and IS FINAL) can be implemented in this way and thus mathematically guaranteed. This means that any network that uses such a PGM as the final activation layer would therefore \textbf{be unable to produce hallucinations of these types}. This approach also has the benefit of allowing the final layer to encode learnt information about which phases are likely to lead into each other separately from this representation. The fact that the sum-product belief propagation algorithm terminates in a finite number of steps also implies the existence and boundedness of its derivatives, allowing it to be used in tandem with any type of loss function, potentially those encouraging the other topological constraints. It also comes equipped with its own version of the Negative Log-Likelihood loss to replace the regular Cross Entropy loss used for softmax during training time.

\section{Simulation Methods}
To illustrate this, we have created a simulation of what surgical workflow recognition in a basic robot-assisted hysterectomy would look like, using Figure \ref{fig:gSPM} as the basis for simulating the order of the different phases. In this simulation, 50\% of the surgeries were performed with the left-side first and the rest with the right-side first. The amount of time in each phase was extracted from a small database of four hysterectomies annotated by one expert gynaecological surgeon annotated at 1 Hz.
For each time point, a vector from a phase-specific 16-dimensional Gaussian distribution is selected. These vectors are then temporally smoothed and then unit standard deviation Gaussian noise is added to mimic the smooth transition between surgical phases as well as to complicate the noise dynamics. The means of said Gaussian distributions are themselves sampled from a unit covariance multivariate Gaussian, giving a slight bias to each label while also overlapping highly. For conceptually similar classes (i.e. the surgical preparation phases and the dissection phases), each mean is moved 75\% of the distance towards their average mean. The covariance matrices were constructed using a common spherical component and a random diagonalized form, $\Sigma = \alpha I + PDP^T$, where $\alpha$ is a constant (1 in our case), $P$ is a uniformly random orthonormal matrix and $D$ is a positive real diagonal matrix $D$ with diagonal elements sampled from a log-normal distribution with $\sigma=1$. These parameters are largely ad-hoc but were designed to be more difficult that the frame-wise features measured by Tayupo \textit{et al.} \cite{tayupo2026please} in their work on automatic surgical phase recognition for hysterectomy in order to generate similar results in terms of the reference softmax/argmax based model. That being said, the simulation is sufficiently distant from real data that it should be seen more as a verification strategy rather than a validation strategy \cite{thacker2004concepts}. The same distributions are used for all data within a single repetition of the experiment, and 200 datapoints are generated with 100 used in training, 100 for testing. A fixed number of 10 epochs is chosen for all simulations and each is sampled at 10 Hz.

The networks learned are simple series of alternating convolution of width 5 timepoints (approximately 10 seconds) and ReLU activations. The depth of the network has been set to three convolution layers each with 32 channels, giving the network the ability to model some temporal evolution. To make the number of network parameters equivalent, only the bias terms are used for the PGM making it exactly equivalent to softmax/argmax in the absence of linear topological predicates. For each repetition, the networks are given the same initialisation and training data presentation order, meaning that they can be directly compared. During testing, multiple predictions for each model are computed using:
\begin{itemize}
    \item no topological information (i.e. argmax);
    \item the same constraints as shown in Figure \ref{fig:gSPM}; or
    \item the ground-truth phase ordering.
\end{itemize}
The former two are to evaluate the differential efficacy of applying these topological constraints in training- vs. application-time whereas the last one is to show potential application for post-operative computer-assisted data annotation. The simulation code, solution algorithms, and results data are provided open-source at \url{https://github.com/JSHBaxter/DeepFlow/tree/master/LTL}. Results are analysed using a series of Wilcoxon paired signed rank tests, pairing by average testing dataset performance across the number of simulations ($N=25$). Statistical significance is considered to be $p < 5\%$ after Bonferroni correction for the 22 tests. (Note that tests for ordering hallucination rates between application gSPM and iSPM are not performed as they are both guaranteed to be zero. Similarly, tests differentiating between training model types in terms of ordering hallucination rates for application-time gSPM and iSPM as well as uniqueness hallucination rates for application-time iSPM activation layers are not performed as they are also guaranteed to be zero. Thus, of the 30 possible tests, 8 are excluded leading to the reported test number.)

\begin{table}[b!]
    \centering
    \begin{tabular}{|p{0.169\textwidth}||c|c|c||c|c|c|}
        \hline
        $(N=25)$& \multicolumn{3}{c||}{\textbf{Trained with no predicates}} & \multicolumn{3}{c|}{\textbf{Trained with predicates}} \\
        \textbf{Testing type} & Acc (\%) & HRO (\#) & HRU (\#) & Acc (\%) & HRO (\#) & HRU (\#)\\
         \hline
        argmax \newline (i.e. no SPM) & $55.4\pm4.3$* & $9.7\pm0.5$* & $7.3\pm0.7$* & $48.6\pm4.5$ & $11.8\pm0.5$ & $9.7\pm0.4$ \\ \hline
        max-product with gSPM & $62.9\pm3.7$*\textdagger & $0.0\pm0.0$\textdagger & $0.5\pm0.2$*\textdagger & $61.3\pm6.2$\textdagger & $0.0\pm0.0$\textdagger & $2.2\pm0.8$\textdagger\\ \hline
        max-product with iSPM & $64.5\pm4.6$\textdagger & $0.0\pm0.0$ & $0.0\pm0.0$\textdagger & $67.6\pm6.5$*\textdagger & $0.0\pm0.0$ & $0.0\pm0.0$\textdagger\\ \hline
    \end{tabular}
    \caption{Accuracy (Acc) and hallucination rate for ordering predicates (HRO) and for uniqueness predicates (HRU) for surgical workflow recognition models trained and evaluated with and without topological information. Statistically significant improvements due to training type or testing type (i.e. row outperforms that immediately above) are indicated by * and \textdagger, respectively with a threshold of $p<0.01$ after Bonferroni correction.}
    \label{tab:quantres}
\end{table}

\section{Simulation Results}
Quantitative results for accuracy are shown in Table \ref{tab:quantres}. From these results, a few basic interpretations can be made.

Firstly, the simulated data had sufficiently different class distributions that both methods managed to learn a classifier that outperformed random chance although difficult enough to not lead to the same levels of accuracy seen in real data \cite{tayupo2026please}. The application of enforcing topological predicates in application-time significantly improved performance by approximately 10\% shows the potential importance of these priors. Interestingly, incorporating more information into these priors did not lead to a further improvement in terms of accuracy. Naturally, the constraints completely removed errors associated with their predicates, notably the ordering predicates.

Interestingly, in the simulation results, there was a minimal difference between training-time methods assuming the same constrained application-time method was employed with the significance flipping from one training method to the other. When the unconstrained argmax is used in application, the decrease in performance for the training with a gSPM is much more significant, implying that the predicates in the training time prevented the underlying network from learning the topological considerations of the problem, likely because they were already enforced by the final activation layer in training-time. Uniqueness constraints were not explicitly enforced in the gSPM (they were implicitly enforced as a logical consequence of the iSPM) but the use of the ordering constraints nevertheless reduced the number of times the uniqueness constraints were broken. This is positive in that it suggests that being able to constrain particular topological properties likely inherently limits which other properties can reasonably be broken.

\section{Discussion}
The simulation, underlying networks, and their results demonstrate the potential applicability of PGMs to guaranteeing certain topological properties in machine learning generated predictions, with some limitations regarding their fidelity and the fact that they are based on a small single-centre dataset and include some \textit{ad-hoc} parameterisation in terms of the simulated signal's intensity. One particularly important limitation is that the simulation is made from the model also used to create the constraint predicates, meaning that the simulation is more of a \textit{verification} of the method rather than a \textit{validation}. This may not be the case in practice in which there may be annotation errors or where the gSPM is overly restrictive and does not strictly allow for all the surgeries in the dataset. In that sense, the simulation results should not be seen as anything more than indicative of a potential role for these frameworks rather than definitive evidence. Even if one takes these results at face value and assume they can be reproduced in real data, there remain three areas of discussion: regulatory considerations, theoretical limitations, and perspectives on foundation models.

\subsection{Regulatory considerations}
One of the primary motivations of this article is to provide a principled way to distinguish between hallucinations and errors in a way that can be more readily incorporated into regulation, i.e. to be quantified and reported. However, it also opens the door to the possibility for certain types of hallucinations to be entirely prevented, constraining medical machine learning architectures at the final activation layer. This approach bridges the gap between the more empirical regulatory approach taken above and a more formal verification approach which are common in critical software systems but difficult to apply to machine learning methods in general \cite{seshia2022toward}.

In addition to encouraging or ensuring formally correct behaviour, there is still the issue of determining what formalisms should be used and in what contexts, especially given the complexity and underdefinedness of the areas of medical image computing most of interest in machine learning research. One approach specific to surgical process modelling and workflow recognition has been the Delphi method \cite{nyangoh2026international} that relies on the synthesis of opinions from a number of surgical experts. However, this approach can be prone to errors and oversimplification if software or machine learning experts are not also involved to explicitly identify edge cases or elucidate underlying assumptions, often in an iterative or dialectical manner \cite{bialy2017software}. Machine learning again complicates this context, as much of the standardisation needs to happen at a basic ontological level \cite{gibaud2018toward,luschi2023semantic}, which itself is necessary to specified in order for data to be annotated consistently and thus machine learning research to commence.

Lastly, there are also some regulatory difficulties adjacent to the \textit{normative/descriptive} mentioned previously. In the case of a set of normative predicates, there is always the possibility that the surgery does not go quite according to plan, meaning that the machine learning system may be constrained to a set of rules that no longer apply. This illustrates an issue in terms of responsibility, as the machine learning algorithm would still be following what it \textit{knows} to be a good protocol and it is the clinician who has deviated. This point is even more difficult for descriptive rules as it may be difficult to enumerate them all for all possible surgeries while still keeping them relatively constrained. In that case, the problem of responsibility may even extend to those who have designed the rule set, well beyond the immediate realm of a single surgeon with a single instance of a machine learning model. Thus, even identifying appropriate predicates for regulatory purposes, is fraught with conceptual and potentially legal difficulties.

\subsection{Theoretical limitations}
This current approach has a number of limitations. The first is that not all linear topological constraints are possible to integrate with the current PGM-based framework. For example, simply stating that a certain phase $A$ must occur in the sequence is a clear constraint, but not one that can be translated into a constraint on an individual \textit{a priori} known time-point or a collection of time-points. For topological rules such as these, other frameworks would need to be incorporated, such as the loss function approach, and those rules would, by consequence, be less strongly enforced and more difficult to regulate.

The other problem is more nuanced. Certain topological rules (such as uniqueness) would lead to the introduction of loops into the PGM. This problem arises from the fact that the belief propagation algorithms used are only known to convergence and be exact for PGM's without loops. Regarding the question of exactitude, we already know from basic experimentation that incorporating constraints via loopy graphs does not yield exact marginals, but they appear to always converge. There is a strong possibility that further mathematical development could guarantee this convergence, but this investigation remains as future work. This work also needs to draw a distinction between training-time and application-time behaviour as one also requires the existence of computable gradients whereas the other does not. Given that training with predicates increases training time and does not (for the moment) appear to improve performance, the necessity of gradient computation becomes more questionable.

\subsection{Perspectives on foundation models}
One of the interesting results is that training without topological predicates yielded equivalent and quite frequently better results, provided that said predicates were enforced in application-time. This may be highly beneficial for the community in terms of training time and amount of data because it suggests that more generalist pre-trained foundation models with minimal adaptation components can still be used while guaranteeing non-hallucination behaviour. This may address the common concern about the reliability of medical vision foundation models applied to new domains \cite{shi2024survey}.

\section{Conclusions}
This article presents a general method for constraining signal segmentation models to more strictly adhere to certain \textit{a priori} specified topological rules. In the context of surgical workflow recognition, these topological rules take the form of linear temporal logic predicates, some of which can be explicitly incorporated into the final activation layer, constraining a machine learning model to adhere to them. This leads to significantly improve application-time performance both in terms of accuracy as well as minimising specific types of errors which could be associated with hallucination in automatic surgical workflow recognition. This has potential impacts on regulation as it suggests pathways for which these errors could be measured as well as prevented in general, although how said rules and constraints are defined is itself an area that would need to be effectively regulated.

\section*{Acknowledgements}
The authors thank Soline Galuret and Elodie Germani for their discussion and feedback on the topic of this article.

\section*{Disclosures of interest}
The authors have no conflicts of interest to disclose.

%
%
%
%
\bibliographystyle{splncs04}
\bibliography{Paper-0007.bib}

@inproceedings{maynez2020faithfulness,
  title={On faithfulness and factuality in abstractive summarization},
  author={Maynez, Joshua and Narayan, Shashi and Bohnet, Bernd and McDonald, Ryan},
  booktitle={Proceedings of the 58th annual meeting of the association for computational linguistics},
  pages={1906--1919},
  year={2020}
}

@inproceedings{rohrbach2018object,
  title={Object hallucination in image captioning},
  author={Rohrbach, Anna and Hendricks, Lisa Anne and Burns, Kaylee and Darrell, Trevor and Saenko, Kate},
  booktitle={Proceedings of the 2018 Conference on Empirical Methods in Natural Language Processing},
  pages={4035--4045},
  year={2018}
}

@article{tayupo2026please,
  title={Please follow the rules: surgical workflow recognition constrained by linear temporal logic},
  author={Tayupo, Dario and Huaulm{\'e}, Arnaud and Nyangoh Timoh, Krystel and Baxter, John SH and Jannin, Pierre},
  journal={International Journal of Computer Assisted Radiology and Surgery},
  pages={1--10},
  year={2026},
  publisher={Springer}
}

@article{liu2025deep,
  title={Deep learning in surgical process modeling: A systematic review of workflow recognition},
  author={Liu, Zhenzhong and Chen, Kelong and Wang, Shuai and Xiao, Yijun and Zhang, Guobin},
  journal={Journal of Biomedical Informatics},
  volume={162},
  pages={104779},
  year={2025},
  publisher={Elsevier}
}

@article{xu2022don,
  title={Don't pour cereal into coffee: Differentiable temporal logic for temporal action segmentation},
  author={Xu, Ziwei and Rawat, Yogesh and Wong, Yongkang and Kankanhalli, Mohan S and Shah, Mubarak},
  journal={Advances in Neural Information Processing Systems},
  volume={35},
  pages={14890--14903},
  year={2022}
}

@article{nyangoh2026international,
  title={International expert consensus-driven surgical process model for robot-assisted hysterectomy: Delphi study results},
  author={Nyangoh Timoh, Krystel and Galuret, Soline and H{\'e}bert, Thomas and Aza{\"\i}s, Henri and Barahona, Marc and Becker, Sven and Bolze, Pierre Adrien and Boisram{\'e}, Thomas and Borghese, Bruno and Carbonnel, Marie and others},
  journal={Surgical Endoscopy},
  volume={40},
  number={1},
  pages={541--552},
  year={2026},
  publisher={Springer}
}

@article{pearl1987distributed,
  title={Distributed revision of composite beliefs},
  author={Pearl, Judea},
  journal={Artificial Intelligence},
  volume={33},
  number={2},
  pages={173--215},
  year={1987},
  publisher={Elsevier}
}

@article{baxter2017directed,
  title={Directed acyclic graph continuous max-flow image segmentation for unconstrained label orderings},
  author={Baxter, John SH and Rajchl, Martin and McLeod, A Jonathan and Yuan, Jing and Peters, Terry M},
  journal={International Journal of Computer Vision},
  volume={123},
  number={3},
  pages={415--434},
  year={2017},
  publisher={Springer}
}

@article{baxter2015shape,
  title={Shape complexes in continuous max-flow hierarchical multi-labeling problems},
  author={Baxter, John SH and Yuan, Jing and Peters, Terry M},
  journal={arXiv preprint arXiv:1510.04706},
  year={2015}
}

@article{seshia2022toward,
  title={Toward verified artificial intelligence},
  author={Seshia, Sanjit A and Sadigh, Dorsa and Sastry, S Shankar},
  journal={Communications of the ACM},
  volume={65},
  number={7},
  pages={46--55},
  year={2022},
  publisher={ACM New York, NY, USA}
}

@incollection{bialy2017software,
  title={Software engineering for model-based development by domain experts},
  author={Bialy, M and Pantelic, V and Jaskolka, J and Schaap, A and Patcas, L and Lawford, M and Wassyng, A},
  booktitle={Handbook of system safety and security},
  pages={39--64},
  year={2017},
  publisher={Elsevier}
}

@article{luschi2023semantic,
  title={Semantic ontologies for complex healthcare structures: a scoping review},
  author={Luschi, Alessio and Petraccone, Camilla and Fico, Giuseppe and Pecchia, Leandro and Iadanza, Ernesto},
  journal={IEEE Access},
  volume={11},
  pages={19228--19246},
  year={2023},
  publisher={IEEE}
}

@article{shi2024survey,
  title={A survey on trustworthiness in foundation models for medical image analysis},
  author={Shi, Congzhen and Rezai, Ryan and Yang, Jiaxi and Dou, Qi and Li, Xiaoxiao},
  journal={arXiv preprint arXiv:2407.15851},
  year={2024}
}

@article{lalys2014surgical,
  title={Surgical process modelling: a review},
  author={Lalys, Florent and Jannin, Pierre},
  journal={International journal of computer assisted radiology and surgery},
  volume={9},
  number={3},
  pages={495--511},
  year={2014},
  publisher={Springer}
}

@article{gibaud2018toward,
  title={Toward a standard ontology of surgical process models},
  author={Gibaud, Bernard and Forestier, Germain and Feldmann, Carolin and Ferrigno, Giancarlo and Gon{\c{c}}alves, Paulo and Haidegger, Tam{\'a}s and Julliard, Chantal and Kati{\'c}, Darko and Kenngott, Hannes and Maier-Hein, Lena and others},
  journal={International journal of computer assisted radiology and surgery},
  volume={13},
  number={9},
  pages={1397--1408},
  year={2018},
  publisher={Springer}
}

@article{thacker2004concepts,
  title={Concepts of model verification and validation},
  author={Thacker, Ben H and Doebling, Scott W and Hemez, Francois M and Anderson, Mark C and Pepin, Jason E and Rodriguez, Edward A},
  year={2004},
  publisher={Los Alamos National Lab., Los Alamos, NM (US)}
}
\end{document}